\documentclass[5p]{elsarticle}

\usepackage{lineno,hyperref}
\modulolinenumbers[5]

\usepackage{prletters}
\usepackage{graphicx}
\usepackage{setspace}
\usepackage{booktabs}
\usepackage{multirow}
\usepackage{float}
\usepackage{mathtools}
\usepackage{soul}
\usepackage{color}
\usepackage{enumitem}
\usepackage{hyperref}
\usepackage{diagbox}
\usepackage{makecell}
\usepackage[toc,page]{appendix}
\usepackage[colorinlistoftodos]{todonotes}
\usepackage{subfigure}
\usepackage{amsmath}
\usepackage{amssymb}
\usepackage{mathrsfs}
\usepackage{bm}
\usepackage[ruled,linesnumbered]{algorithm2e}

\begin{document}

\begin{frontmatter}

\title{MTHetGNN: A Heterogeneous Graph Embedding Framework for Multivariate Time Series Forecasting}

\fntext[eq]{Equal contribution.}
\cortext[cor]{Corresponding author.}

\author[mymainaddress]{Yueyang Wang\corref{cor}\fnref{eq}}
\ead{yueyangw@cqu.edu.cn}

\author[mymainaddress,mysecondaryaddress]{Ziheng Duan\fnref{eq}}

\author[mysecondaryaddress]{Yida Huang}
\author[mysecondaryaddress]{Haoyan Xu}
\author[mysecondaryaddress]{Jie Feng}
\author[mysecondaryaddress]{Anni Ren}

\address[mymainaddress]{School of Big Data and Software Engineering, Chongqing University, Chongqing, 401331, China}
\address[mysecondaryaddress]{College of Control Science and Engineering, Zhejiang University, Zhejiang, 310027, China}

\begin{abstract}

Multivariate time series forecasting, which analyzes historical time series to predict future trends, can effectively help decision-making. Complex relations among variables in MTS, including static, dynamic, predictable, and latent relations, have made it possible to mining more features of MTS. Modeling complex relations are not only essential in characterizing latent dependency as well as modeling temporal dependence, but also brings great challenges in the MTS forecasting task. However, existing methods mainly focus on modeling certain relations among MTS variables. In this paper, we propose a novel end-to-end deep learning model, termed \underline{M}ultivariate \underline{T}ime Series Forecasting via \underline{Het}erogeneous \underline{G}raph \underline{N}eural \underline{N}etworks (MTHetGNN). To characterize complex relations among variables, a \emph{relation embedding} module is designed in MTHetGNN, where each variable is regarded as a graph node, and each type of edge represents a specific static or dynamic relationship. Meanwhile, a \emph{temporal embedding} module is introduced for time series features extraction, where involving convolutional neural network (CNN) filters with different perception scales. Finally, a \emph{heterogeneous graph embedding} module is adopted to handle the complex structural information generated by the two modules. Three benchmark datasets from the real world are used to evaluate the proposed MTHetGNN. The comprehensive experiments show that MTHetGNN achieves state-of-the-art results in the MTS forecasting task.
\end{abstract}


\end{frontmatter}


\section{Introduction}

In real-world scenarios, data can be naturally expressed as multivariate time series (MTS), which are generally composed of multiple single-dimensional time series of the same object, such as the observation of the same object by multiple sensors, the traffic flow of each block in the same area, or the exchange rate information of different countries \cite{xu2020parallel}.
Time series analysis, which analyzes historical time series and gets predictions about future trends, has been proved to be highly effective in making helpful strategic decisions and received increasing attention in recent years. However, most time series analysis approaches mainly focus on capturing a specific relation among variables and may not handle MTS efficiently. 

To provide probabilistic explanations for more reasonable predictions, the core of MTS forecasting is to make full use of the following two significant characteristics: (1) the internal temporal dependency pattern of each single-dimensional time series; (2) the rich spatial relations among different variables in MTS. On the one hand, the variables in each time series depend on its historical values. For instance, the activity of the sun shows a periodic pattern in historical observations, the t-th time value of the variable may be similar to one historical value.
On the other hand, regarding each time series in MTS as a variable, the interdependency among variables is useful to exploit.
For example, the future traffic flow of a specific street is easier to be predicted by introducing the traffic information of neighboring areas, while the impact from the region farther away is relatively slight. Therefore, considering these internal and external relations of MTS can be an effective guideline for forecasting. Besides, apart from above interdependency information which is available and helpful for MTS forecasting, there also exists relations among variables that are unknown or changing over time, implicitly exhibited. However, there is a limitation on the existing methods to exploit latent and rich interdependencies among variables efficiently and effectively.

Over the years, researchers have adopted different techniques and assumptions to model MTS. 
Classical MTS forecasting models consider statistic information of historical measurement and make the prediction.
Autoregressive integrated moving average model (ARIMA) is a popular machine learning model which can be applied flexibly to various types of time series with a high computational cost.
VAR \cite{zhang2003time} extends the autoregressive (AR) model to multivariate time series, thus it cannot integrate the relations among time series variables.
Many deep learning models, like LSTNet \cite{lai2018modeling} and MLCNN \cite{cheng2019better}, consider the long-term dependency and short-term variance of time series, while they cannot explicitly model the pairwise dependencies among variables. 
Recently, researches found it promising to model multivariate time series using graph neural networks \cite{xu2020multivariate,wu2020connecting,Duan2020multivariate}. 
Time series variables can be considered as nodes in the graph while the interrelations among them as edges.
The information of MTS is stored in this graph structure and is then processed by the following graph neural networks.
However, TEGNN \cite{xu2020multivariate}, MTGNN \cite{wu2020connecting} can only reveal one type of relation, lacking the ability to model both static and dynamic relations in time series. 
The classical machine learning methods and mentioned deep learning methods can not fully explore the implicit relations among time series.

In this work, a novel framework, termed \emph{\underline{M}}ultivariate \emph{\underline{T}}ime Series Forecasting with \emph{\underline{Het}}erogeneous \emph{\underline{G}}raph \emph{\underline{N}}eural \emph{\underline{N}}etwork (MTHetGNN) is proposed and applied for the MTS forecasting task. MTHetGNN embeds each relation or interdependency into each graph structure and fuses all graph structures with temporal features. The relation embedding module considers both static, prior, and dynamic, latent relations among variables, characterizing the global relations (such as similarity and causality) and dynamic local relations among time series, respectively. In addition, convolution neural networks (CNN) are used for temporal feature extraction. Finally, heterogeneous graph neural networks take the output of the former modules and tackle the complex structural embedding of graph structure generated by MTS for the forecasting task. Thus our major contributions are:
\begin{itemize}
\item We first propose a heterogeneous graph network-based framework that is compatible with taking full advantage of rich relations among variables of MTS.
\item We construct a relation embedding module to explore the relations among time series in both dynamic and static approaches. 
\item We conduct extensive experiments on MTS benchmark datasets. The experimental results validate that the performance of the proposed method is better than state-of-the-art models.
\end{itemize}

\section{Related Work}
\subsection{Multivariate Time Series Forecasting}
So far, there have been many deep learning models proposed on time series forecasting.
They use classic neural network structures to extract feature of time series, such as recurrent neural network (RNN) \cite{elman1990finding}. 
On the basis of these units, scholars have designed many improved frameworks to make them better adapted to time series forecasting tasks. Lai et al. propose the LSTNet framework \cite{lai2018modeling}, which uses CNN and RNN-skip component to capture the long-term and short-term patterns of MTS, respectively. 
Cheng et al. \cite{cheng2020towards} build a MLCNN framework based on LSTM and CNN for fusing near and distant future visions. 
To deal with the limitations of RNN and temporal convolution networks (TCNs), Cirstea et al. \cite{cirstea2021enhancenet} propose a framework EnhanceNet to capture both distinct temporal dynamics and dynamic entity correlations.


\subsection{Graph Neural Network}
Nowadays, neural networks have been employed for representing graph structured data \cite{xu2021graph,xu2020cosimgnn}, such as social networks and knowledge bases. 
Originated from Graph Signal Processing \cite{ortega2018graph}, classical convolutions are extended to spectral domain, which is space and time consuming.
Further research \cite{defferrard2016convolutional} approximate the spectral convolution using K-hops polynomials, reducing the time complexity effectively. Finally, GCN \cite{kipf2016semi}, a scalable approach chose the convolutional architecture via a localized approximation with Chebyshef Polynomial, which is an efficient variant and can operate on graphs directly. 
However, these methods can only implement on undirected graphs. Previously in form of recurrent neural networks, Graph Neural Networks (GNNs) are proposed to operate on directed graphs.

\subsection{Heterogeneous Network Embedding}

Conventional methods for dealing with heterogeneous networks usually start with the extraction of typed structural features, aiming to pursue meaningful vector representations for each node for downstream applications \cite{duan2021connecting}. 
However, this task needs to consider structural information composed of multiple types of nodes and edges, which is challenging. 
Many methods of dealing with heterogeneous networks involve the concept of \textit{meta-structure} \cite{dong2020heterogeneous}. 
For example, metapath2vec \cite{dong2017metapath2vec} uses a path composed of nodes obtained from random walks guided by metapaths, and considers heterogeneous semantic information to model the context of nodes. 
NSHE \cite{zhao2020network} proposed some delicate designs, e.g., network schema sampling and multi-task learning, which preserves a high-order structure in heterogeneous networks and alleviates the meta-path selection dilemma in meta-path-guided heterogeneous network embeddings.

The heterogeneity and rich semantic information bring significant challenges for designing heterogeneous graph neural networks. 
At the same time, the attention mechanism shows exciting advancements in deep learning.
On this basis, some researchers \cite{Duan2019heterogeneous} have applied the attention mechanism to heterogeneous graphs, showing excellent results.
For example, HAN \cite{wang2019heterogeneous} uses metapath to model higher-order similarities (not directly using first-order neighbors), and uses the attention mechanism to learn different weights for different neighbors.
HGAT \cite {yang2021hgat} use a dual-level attention mechanism, including node-level and type-level attention. 
Specifically, the former aims to learn the importance of nodes and their neighbors based on meta-paths, while the latter can learn the importance of different meta-paths. 
The dual-level attention mechanism can fully consider the rich information in heterogeneous graphs.
Many attention variants in heterogeneous networks were also proposed.
For instance, \cite{zhang2020sr} proposed a new heterogeneous graph attention network based on high-order symmetric relations (SR-HGAT).
SR-HGAT considers the characteristics of nodes and high-order relations simultaneously and uses an aggregator based on a two-layer attention mechanism to capture basic semantics effectively.
And \cite{carletti2021predicting} used a relation-wise Graph Attention Network with a Relation Attention Module (RAM) to predicting polypharmacy side effects.

\section{The Framework}
\subsection{Task Formulation}
In this work, we explore the task of multivariate time series forecasting. Formally, given a time series $X_i = \{x_{i1}, x_{i2}, ..., x_{iT}\}$, where $x_{it}\in \mathcal{R}^n$ is the observation with $n$ variables at time stamp $t$ from the $i^{th}$ sample. $T$ is the number of time stamps. 
The task is to predict the future value $ x_{t+h}$ where $h$ denotes the horizon ahead of the current time stamp. Considering all samples $\mathcal{X} = \{X_1,X_2, ..., X_s\}$ where $s$ is the number of samples, and the ground truth forecasting value $\mathcal{Y} = \{Y_1,Y_2, ..., Y_s\}$, we aims to model the mapping from $\mathcal{X}$ to $\mathcal{Y}$. 

\begin{figure*}[t]
\centering
\includegraphics[width=1\linewidth]{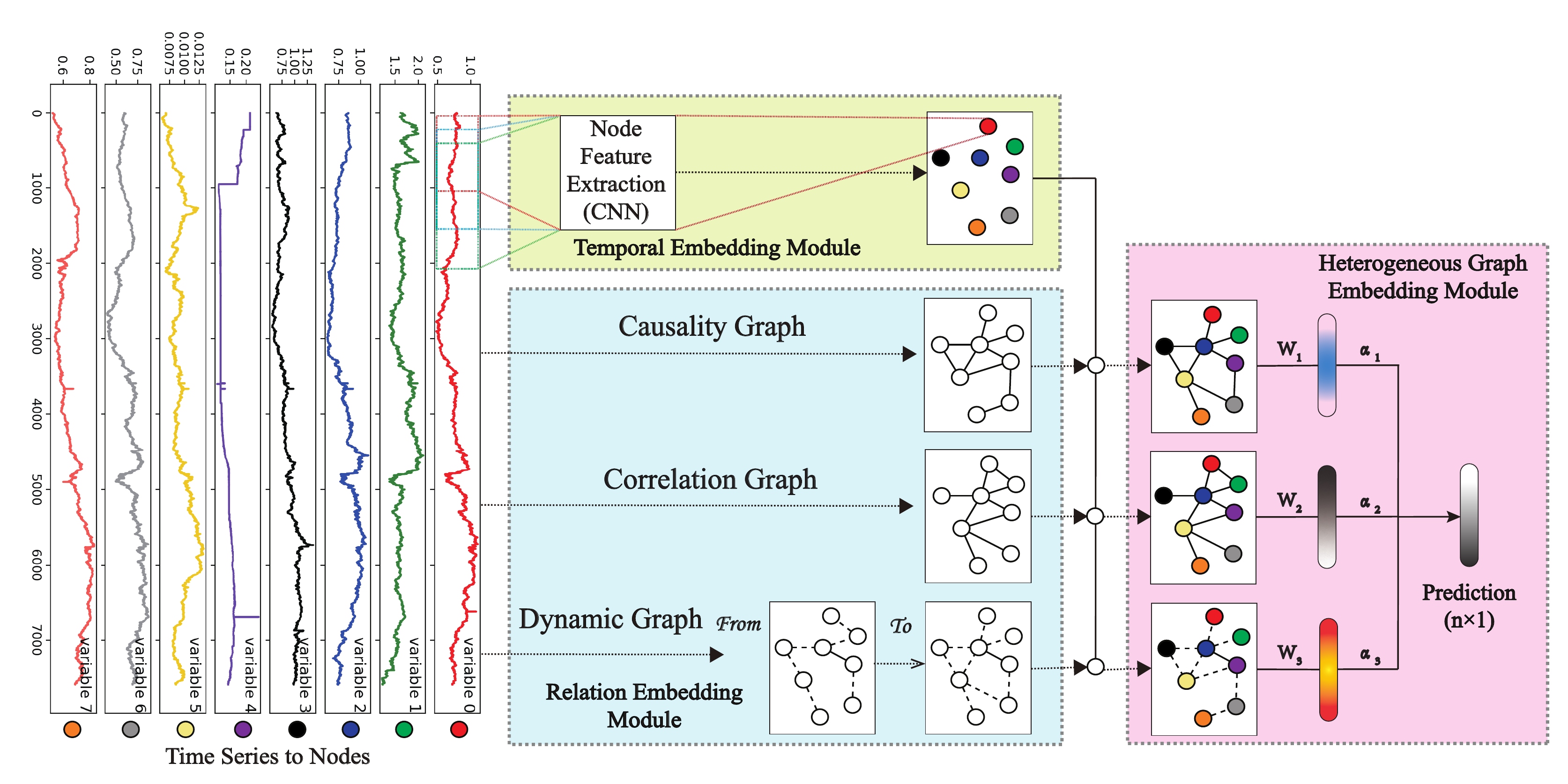}
\vspace{-6mm}
\caption{The general architecture of our model MTHetGNN, which contains three modules and they jointly learn in an end-to-end fashion.
Each time series is converted into a node in the graph (the color of the time series is consistent with the color of the node).
Temporal embedding module captures temporal features as node features.
Relation embedding module captures static, prior and dynamic, latent spatial relations among variables.
The "From" and "To" in the figure indicate that the nodes' relations are updated from the state at the previous moment to the state at this moment. 
Heterogeneous graph embedding module exploits and fuses rich spatial patterns with temporal features for better forecasting.}
\label{fig:1}
\vspace{-3mm}
\end{figure*}

\subsection{Model Architecture}
We first narrate our proposed model MTHetGNN, i.e., \emph{\underline{M}}ultivariate \emph{\underline{T}}ime Series Forecasting via \emph{\underline{Het}}erogenous \emph{\underline{G}}raph \emph{\underline{N}}eural \emph{\underline{N}}etworks in detail, which is a framework for modeling multivariate time series from a graph perspective with compatible modules. An overview of MTHetGNN is illustrated in Figure \ref{fig:1}. MTHetGNN contains three components: \textit{Temporal Embedding Module}, \textit{Relation Embedding Module} and \textit{Heterogeneous Graph Embedding Module}. To capture temporal features from time series, we adopt CNN with multi receptive fields in temporal embedding module, which could be replaced by methods like RNN and its variants. The relation embedding module captures different internal static and dynamic relations among variables in MTS. 
Taking the above two modules' output, heterogeneous graph embedding models can exploit rich spatial dependencies in graph structures to model heterogeneity in time series. The three modules jointly learn in an end-to-end fashion to exploit and fuse priori information, dynamic latent relations and temporal features. 

\subsection{Temporal Embedding Module}
This module aims to capture temporal features by applying multiple CNN filters. As shown in Figure \ref{fig:1}, CNN filters with different receptive fields are applied on multivariate time series, thus features under different periods are extracted from raw signals.
Follow the concept of \textit{inception} suggested in \cite{szegedy2015going}, we use $p$ CNN filters with kernel sizes $(1\times k_i)(i=1, 2, 3, ..., p)$ to scan through input time series $x$ to capture features at multiple time scales. Here set of convolution filters with kernel sizes of $[1\times3, 1\times5, 1\times7]$ are used to capture features at multiple time scales. 

\subsection{Relation Embedding Module}
The relation embedding module learns graph adjacency matrix to model the internal relations among time series. 

We model implicit relations in MTS variables using both static and dynamic strategies. From the static perspective, we use correlation coefficient and transfer entropy to model static linear relationships and implicit causality among variables. From the dynamic perspective, we adopt dynamic graph learning concept and learn the graph structure adaptively, modeling time-varying graph structure. By using the above three strategies, varies adjacency matrices are generated and then fed into heterogeneous graph neural networks to interpret the relations between graph nodes in both static and dynamic way.

Recall that for all samples $\mathcal{X} = \{X_1,X_2, ..., X_s\}$, the similarity adjacency matrix $A^{sim}\in \mathcal{R}^{n\times n}$ is generated by:
\begin{equation}
    A^{sim} = Similarity(\mathcal{X}),
\end{equation}
where $Similarity$ is a distance metric which measures the pairwise similarity scores between time series. Existed work to measure distance include \emph{Euclidean Distance}, \emph{Landmark Similarity} and \emph{Dynamic Time Warping (DTW)}, etc. Here we adopt \emph{correlation} and \emph{coefficient} to measure the global correlation among time series, offering a priori knowledge of overal linear relation. Thus element in $A^{sim}$ is generated by:
\begin{equation}
    A_{ij}^{sim} = \frac{Cov(X_i,X_j)}{\sqrt{D(X_i)}\sqrt{D(X_j)}},
\end{equation}
where $Cov(X_i,X_j)$ is the covariance between $X_i$ and $X_j$, $D(X_i)$ and $D(X_j)$ is the variance of time series $X_i$ and $X_j$ respectively.

The causality adjacency matrix $A^{cas}\in \mathcal{R}^{n\times n}$ is generated by:
\begin{equation}
    A^{cas} = Causality(\mathcal{X}),
\end{equation}
where $Causality$ is a metric to measure causality between time series. Various efforts have been made to measure causal inference among variables, such as Granger, etc. Here we use Transfer Entropy (TE) to process non-stationary time series, a pseudo-regression relationship will be produced in which pairwise time series be considered causal if they have an overall trend caused by common factors. The causality mentioned here is not strict, but the value is helpful for predicting. Given graph variables $X$ and $Y$, the transfer entropy of variable $A$ to $B$ is defined as:
\begin{align}
    T_{B\rightarrow A}&= H(A_{t+1}|A_t)-H(A_{t+1}|A_t,B_t),
\end{align}
in which the concept of conditional entropy is used. Let $m_t$ represent the value of variable $m$ at time $t$, and $m_t^{(k)}=[m_t,m_{t-1}, ..., m_{t-k+1}]$. $H(M_{t+1}|M_t)$ is the conditional entropy representing the information amount of $M_{t+1}$ under the condition that the variable $M_t$ is known:
\begin{equation}
    H(M_{t+1}|M_t) =\sum p\left(m_{t+1},m_t^{(k)}\right)\log_2p\left(m_{t+1}|m_t^{(k)}\right).
\end{equation}
By calculate transfer entropy between time series, the element in causality adjacency matrix $A^{cas}$ is calculated as:

\begin{equation}
    A_{ij}^{cas} = T_{X_i\rightarrow X_j} - T_{X_j\rightarrow X_i}.
\end{equation}

The third strategy adopts the concept of dynamic evolving networks \cite{skarding2020foundations}. 
In a certain period of time, the time series persist to establish a relatively stable graph network, and node properties like node degree can be updated in training process. 
We propose a dynamic relation embedding strategy, which learns the adjacency matrix $A^{DA}\in \mathcal{R}^{n\times n}$ adaptively to model latent relations in the given time series sample $X_i$, denoted as:

\begin{equation}
    A^{dy} = Evolve(X_i).
\end{equation}

Given the input time series $X_k = {x_1, x_2, ..., x_n} \in \mathcal{R}^{n\times T}$ from the $k_{th}$ sample with length $T$, where $x_i,x_j$ denote the $i^{th}, j^{th}$ time series. We first calculate the distance matrix $D$ between sampled time series:
\begin{equation}
    D_{ij}= \frac{exp(-\sigma(distance(x_i,x_j)))}{\sum_{p = 0}^n \sigma(exp(-\sigma(distance(x_i,x_p)))},
\end{equation}
where $distance$ denotes the distance metric.
The dynamic adjacency matrix $A^{dy}\in \mathcal{R}^{n\times n}$ can be calculated as:
\begin{align}
    A^{dy} &= \sigma(D W).
\end{align}
$W$ is a learnable parameter and $\sigma$ is an activation function.

Normalization is applied to the output of each strategy respectively to form three normalized adjacency matrix. What's more, to improve training efficiency, reduce the effect of noise, amplify the effective relations and make the model more robust, threshold is set to make all the adjacency matrices sparse:

\begin{equation}
A_{ij}^r=\left\{
\begin{aligned}
    A_{ij}^r &  & A_{ij}^r >= threshold\\
    0 &  & A_{ij}^r < threshold
\end{aligned}
\right.
\end{equation}

\subsection{Heterogeneous Graph Embedding Module}
This module could be viewed as a graph based aggregation method, which fuses temporal features and spatial relations between time series to get forecasting results. 

Our model is primarily motivated by rGCNs \cite{schlichtkrull2018modeling} which learns an aggregation function that is representation-invariant and can operate on large-scale relational data. We adopt the idea of fusing node embeddings of each heterogeneous graph with attention mechanism. We propose the following propagation function:  
\begin{equation}
    H^{(l+1)}=\sigma \left( H^{(l)} W_0^{l} + \sum\limits_{r\in\mathcal{R}}{softmax(\alpha_r) \hat A_r H^{(l)}W_r^{(l)}} \right),
\end{equation}
where $H^{(l)}$ is the matrix of node embedding in the $l^{th}$ layer, $H^{(0)}=X$; $\alpha_{(r)}$ is the weight coefficient of each heterogeneous graph, and $softmax(\alpha_r) = \frac{exp(\alpha_r)}{\sum_{i = 1}^{|R|}exp(\alpha_i)}$. $W_o^{(l)}$ and $W_r^{(l)}$ are layer-specific weight matrix. $\sigma$ is a nonlinear activation function, usually being $Relu$.
Inspired by scaled dot-product attention mechanism \cite{vaswani2017attention}, we extend the use of the dot-product operation to compute the attention coefficients between heterogeneous graph neural network.

\section{Objective Function}
$\mathcal{L}_2$ loss is used in many forecasting tasks:
\begin{equation}
    minimize(\mathcal{L}_2) = \frac{1}{k}\sum\limits_i^k\sum\limits_j^n(y_{ij}-\hat y_{ij})^2,
\end{equation}
where $k$ is the training size and $n$ is the variables in time series. $\hat y$ is the prediction and $y$ is the ground truth.
Researchers have found that objective function using $\mathcal{L}_1$ loss has a stable gradient for different inputs, which can reduce the impact of outliers while avoiding gradient explosions:
\begin{equation}
    minimize(\mathcal{L}_1) = \frac{1}{k}\sum\limits_i^k\sum\limits_j^n|y_{ij}-\hat y_{ij}|
\end{equation}

We use the Adam optimizer and decide which objective function to use by the performance on the validation set.

\section{Experiments}
We conduct experiments on MTHetGNN model on three benchmark datasets and compare the performance of MTHetGNN with eight baseline methods for multivariate time series forecasting tasks.
\subsection{Data}
We use three benchmark datasets which are commonly used in MTS forecasting.
The details are as following:
\begin{itemize}
    \item Exchange-Rate: The exchange rate data from eight countries, including UK, Japan, New Zealand, Canada, Switzerland, Singapore, Australia and China, ranging from 1990 to 2016.
    \item Traffic\footnote{http://pems.dot.ca.gov}: The traffic highway occupancy rates measured by 862 sensors in San Francisco from 2015 to 2016 by California Department of Transportation.
    \item Solar-Energy\footnote{http://www.nrel.gov/grid/solar-power-data.html}: Continuous collected Solar energy data from the National Renewable Energy Laboratory, which contains the solar energy output collected from 137 photovoltaic power plants in Alabama in 2007.
\end{itemize}

\subsection{Methods for Comparison}
We evaluate the performance of MTHetGNN with other nine baseline models on MTS forecasting task.
The overview of baseline methods are summarized as bellow:
\begin{itemize}
    \item VAR-MLP \cite{zhang2003time}: A machine learning model, which is the combination of Multilayer Perception (MLP) and Autoregressive model.
    \item RNN-GRU \cite{dey2017gate}: A Recurrent Neural Network adopting GRU cell.
    \item LSTNet \cite{lai2018modeling}: A deep learning method, which uses Convolution Neural Network and Recurrent Neural Network to discover both short and long term patterns for time series.
    \item DCRNN \cite{li2017diffusion}: A diffusion convolutional recurrent neural network, which combines graph convolution networks with recurrent neural networks in an encoder-decoder manner. 
    \item Graph WaveNet \cite{wu2019graph}: A spatial-temporal graph convolutional network, which integrates diffusion graph convolutions with 1D dilated convolutions.
    \item EnhanceNet \cite{cirstea2021enhancenet}: An enhanced framework, which integrates Distinct Filter Generation Network and Dynamic Adjacency Matrix Generation Network to boost the forecasting accuracy.
    \item MLCNN \cite{cheng2019better}: A deep neural network which fuses forecasting information of different future time.
    \item MTGNN \cite{wu2020connecting}: A graph neural network designed for multivariate time series forecasting.
    \item TEGNN \cite{xu2020multivariate}: A novel deep learning framework to tackle forecasting problem of graph structure generated by MTS considering causal relevancy. 
\end{itemize}

\subsection{Metrics}
Conventional evaluation metrics are used to evaluate all methods: \textit{Relative Squared Error (RSE)},  \textit{Relative Absolute Error (RAE)}, and \textit{Empirical Correlation Coefficient (CORR)}.
For \textit{RSE} and \textit{RAE}, lower value is better, while for \textit{CORR}, higher value is better.


\begin{table*}

    \caption{MTS forecasting results measured by RSE/RAE/CORR score over three datasets. The best performance results are bolded.}
    \centering
    \scalebox{0.68}{
    \begin{tabular}{lc|cccc|cccc|cccc} 
    \toprule
    Dataset&&\multicolumn{4}{c|}{Exchange-Rate}& \multicolumn{4}{c|}{Solar} & \multicolumn{4}{c}{Traffic} \\
    \midrule
    &&\multicolumn{1}{c}{horizon}&\multicolumn{1}{c}{horizon}&\multicolumn{1}{c}{horizon}&\multicolumn{1}{c|}{horizon}
    
    &\multicolumn{1}{c}{horizon}&\multicolumn{1}{c}{horizon}&\multicolumn{1}{c}{horizon}&\multicolumn{1}{c|}{horizon}
    
    &\multicolumn{1}{c}{horizon}&\multicolumn{1}{c}{horizon}&\multicolumn{1}{c}{horizon}&\multicolumn{1}{c}{horizon}\\
    
    Methods&Metrics
    &\multicolumn{1}{c}{3} &\multicolumn{1}{c}{6} &\multicolumn{1}{c}{12} &\multicolumn{1}{c|}{24} 
    &\multicolumn{1}{c}{3} &\multicolumn{1}{c}{6} &\multicolumn{1}{c}{12} &\multicolumn{1}{c|}{24} 
    &\multicolumn{1}{c}{3} &\multicolumn{1}{c}{6} &\multicolumn{1}{c}{12} &\multicolumn{1}{c}{24}\\
    \midrule
    
    \multirow{5}{*}{\textsc{VAR}}
        &RSE& 0.0186 & 0.0262 & 0.0370 & 0.0505 & 0.1932 & 0.2721 & 0.4307 & 0.8216& 0.5513 & 0.6155 & 0.6240 & 0.6107\\
        &RAE& 0.0141 & 0.0208 & 0.0299 & 0.0427 & 0.0995 & 0.1484 & 0.2372 & 0.4810 & 0.3909 & 0.4066 & 0.4177 & 0.4032\\
        &CORR& 0.9674 & 0.9590 & 0.9407 & 0.9085& 0.9819 & 0.9544 & 0.9010 & 0.7723 & 0.8213 & 0.7826 & 0.7750 & 0.7858\\
    \midrule
    
    \multirow{5}{*}{\textsc{RNN-GRU}}
        &RSE& 0.0200 & 0.0262 &0.0366 & 0.0527 & 0.1909 & 0.2686 & 0.4270 & 0.4938 & 0.5200 & 0.5201 & 0.5320 & 0.5428\\
        &RAE& 0.0157 & 0.0209 & 0.0298 & 0.0442 & 0.0946 & 0.1432 & 0.2302 & 0.2849 & 0.3625 & 0.3708 & 0.3669 & 0.3844\\
        &CORR& 0.9772 & 0.9688 & 0.9534 & 0.9272 & 0.9832 & 0.9660 & 0.9112 & 0.8808 & 0.8436 & 0.8459 & 0.8316 & 0.8232\\
    \midrule
    
    \multirow{5}{*}{\textsc{LSTNET}}
        &RSE& 0.0216 & 0.0277 & 0.0359 & 0.0482 & 0.1940 & 0.2755 & 0.4332 & 0.4901 & 0.4769 & 0.4890 & 0.5110 & 0.5037\\
        &RAE& 0.0171 & 0.0226 & 0.0295 & 0.0404 & 0.0999 & 0.1510 & 0.2413 & 0.2997 & 0.3161 & 0.3291 & 0.3435 & 0.3441\\
        &CORR& 0.9749 & 0.9678 & 0.9534 & 0.9353 & 0.9825 & 0.9633 & 0.9065 & 0.8673 & 0.8730 & 0.8657 & 0.8534 & 0.8537\\
    \midrule
    
    \multirow{5}{*}{\textsc{DCRNN}}
        &RSE& 0.0197 & 0.0258 & 0.0355 & 0.0485 & 0.1875 & 0.2611 & 0.3661 & 0.4780 & 0.4880 & 0.4957 & 0.5078 & 0.5125\\
        &RAE& 0.0159 & 0.0215 & 0.0291 & 0.0387 & 0.0965 & 0.1475 & 0.1891 & 0.2775 & 0.2938 & 0.3037 & 0.3219 & 0.3395\\
        &CORR& 0.9767 & 0.9699 & 0.9540 & 0.9350 & 0.9817 & 0.9618 & 0.9437 & 0.8611 & 0.8659 & 0.8518 & 0.8398 & 0.8267\\
    \midrule
    
    \multirow{5}{*}{\textsc{Graph WaveNet}}
        &RSE& 0.0189 & 0.0249 & 0.0337 & 0.0457 & 0.1788 & 0.2548 & 0.3315 & 0.4231 & 0.4571 & 0.4538 & 0.4829 & 0.5007\\
        &RAE& 0.0151 & 0.0198 & 0.0279 & 0.0371 & 0.0873 & 0.1257 & 0.1715 & 0.2485 & 0.2755 & 0.2895 & 0.3018 & 0.3111\\
        &CORR& 0.9787 & 0.9715 & 0.9548 & 0.9385 & 0.9839 & 0.9698 & 0.9508 & 0.8798 & 0.8845 & 0.8607 & 0.8585 & 0.8436\\
    \midrule
    
    \multirow{5}{*}{\textsc{EnhanceNet}}
        &RSE& 0.0181 & 0.0244 & 0.0342 & 0.0461 & 0.1791 & 0.2431 & 0.3275 & 0.4028 & 0.4321 & 0.4497 & 0.4751 & 0.4877\\
        &RAE& 0.0144 & 0.0193 & 0.0281 & 0.0384 & 0.0855 & 0.1221 & 0.1689 & 0.2444 & 0.2587 & 0.2699 & 0.2913 & 0.3018\\
        &CORR& 0.9798 & 0.9733 & 0.9582 & 0.9388 & 0.9842 & 0.9655 & 0.9533 & 0.8910 & 0.8891 & 0.8714 & 0.8641 & 0.8569\\
    \midrule
    
    \multirow{5}{*}{\textsc{MLCNN}}
        &RSE& \textbf{0.0172} & 0.0449 & 0.0519 & 0.0438 & 0.1794 & 0.2983 & 0.3673 & 0.5191 & 0.4924 & 0.4992 & 0.5214 & 0.5353\\
        &RAE& \textbf{0.0129} & 0.0334 & 0.0422 & 0.0375 & 0.0844 & 0.1342 & 0.1873 & 0.3131 & 0.3376 & 0.3243 & 0.3766 & 0.3825\\
        &CORR& 0.9780 & 0.9610 & 0.9550 & 0.9407 & 0.9814 & 0.9642 & 0.9210 & 0.8513 & 0.8629 & 0.8416 & 0.8320 & 0.8255\\
    \midrule
    
    \multirow{5}{*}{\textsc{MTGNN}}
        &RSE& 0.0194 & 0.0253 & 0.0345 & 0.0447 & 0.1767 & 0.2342 & 0.3088 & 0.4352 & 0.4178 & 0.4774& 0.4461 & 0.4535\\
        &RAE& 0.0156 & 0.0206 & 0.0283 & 0.0376 & 0.0837 & 0.1171 & 0.1627 & 0.2563 & 0.2435 & 0.2670 & 0.2739 & \textbf{0.2651}\\
        &CORR& 0.9782 & 0.9711 & 0.9564 & 0.9370 & 0.9852 & 0.9727 & 0.9511 & 0.8931 & 0.8960 & 0.8665 & 0.8794 & \textbf{0.8810}\\
    \midrule
    
    \multirow{5}{*}{\textsc{TEGNN}}
        &RSE& 0.0178 & 0.0245 & 0.0363 & 0.0449 & 0.1824 & 0.2612 & 0.3289 & 0.4733& 0.4421 & 0.4433& 0.4508 & 0.4692\\
        &RAE& 0.0135 & 0.0195 & 0.0306 & 0.0388 & 0.0851 & 0.1312 & 0.1766 & 0.2821& 0.2651 & 0.2616 & 0.2740 & 0.2855\\
        &CORR& 0.9815 & 0.9732 & 0.9566 & 0.9352 & 0.9847 & 0.9676 & 0.9379 & 0.8854 & 0.8853 & 0.8820 & 0.8743 & 0.8671\\
    \toprule[1.2pt]
    
    \multirow{5}{*}{\textsc{MTHetGNN}}
        &RSE& 0.0173 & \textbf{0.0238} & \textbf{0.0327} & \textbf{0.0430} & \textbf{0.1668} & \textbf{0.2175} & \textbf{0.2872} & \textbf{0.3862} & \textbf{0.4142} & \textbf{0.4303} & \textbf{0.4376} & \textbf{0.4500}\\
        &RAE& 0.0132 & \textbf{0.0190} & \textbf{0.0266} & \textbf{0.0361} & \textbf{0.0788} & \textbf{0.1111} & \textbf{0.1514} & \textbf{0.2217} & \textbf{0.2349} & \textbf{0.2490} & \textbf{0.2592} & 0.2661 \\
        &CORR& \textbf{0.9824} & \textbf{0.9746} & \textbf{0.9604} & \textbf{0.9415} & \textbf{0.9872} & \textbf{0.9772} & \textbf{0.9583} & \textbf{0.9210} & \textbf{0.8975} & \textbf{0.8887} & \textbf{0.8828} & 0.8776\\
    \midrule
    \label{MainResults}
    
    \end{tabular}}
    \vspace{-0.4cm}
    \label{sec:table1}
    \end{table*}

\subsection{Experimental Details}
On three benchmark datasets, data are split into training set, validation set and testing set in a ratio of $6:2:2$, then we use the model with the best performance based on \textit{RSE}, \textit{RAE} and \textit{CORR} metrics on validation set for testing. 
We conduct grid search strategy over adjustable hyper-parameters for all methods. 
The window size $T$ is set to 32 for all methods. 
For RNN, the hidden RNN layer is chosen from $\{10, 20, 50, 100\}$, the dropout rate is chosen from $\{0.1, 0.2, 0.3\}$.
For LSTNet, the hidden CNN and RNN layer is chosen from $\{20, 50, 100, 200\}$, the length of recurrent-skip is set to $24$.
For DCRNN, both encoder and decoder contain two recurrent layers and each layer has 64 units.
For Graph WaveNet, to cover the input sequence length (or the window size 32), we use 16 layers of Graph WaveNet with a repeat sequence of dilation factors 1, 2, 4, 8.
For EnhanceNet, we use enhanced RNN to extract feature.
The memory dimension and the RNN units are all set to 16.
For MTGNN, the mix-hop propagation depth is set to 2, the activation saturation rate of graph learning layer is set to 3.  
For TEGNN and MTHetGNN, the hidden graph convolutional networks is chosen from $ \{5,10,15, ..., 100\}$.

\subsection{Effectiveness}

Table \ref{sec:table1} summarizes the evaluation results of MTHetGNN and other baseline methods on three benchmark datasets under different settings. Following the settings of LSTNet \cite{lai2018modeling}, we test the model performance on forecasting future values $\{X_{t+3},X_{t+6},X_{t+12},X_{t+24} \}$, which means future value from $3$ to $24$ days over the Exchange-Rate data, $30$ to $240$ minutes over the Solar-Energy data, and $3$ to $24$ hours over the Traffic data, $s$ thus $horizon$ are set to $\{3,6,12,24\}$ for three benchmark datasets respectively. 
As shown in table \ref{sec:table1}, the best results under 4 different $horizon$ with 3 evaluation metrics are set bold.
On the Exchange-Rate dataset, when the horizon is 3, the performance of MLCNN is a little bit better than MTHetGNN; on the Traffic dataset, when the horizon is 24, the performance of MTGNN is slightly better than MTHetGNN.
In other cases, MTHetGNN has better results under all metrics than methods such as DCRNN, Graph WaveNet and their enhanced version EnhanceNet.

TEGNN, MTGNN and MTHetGNN use graph structure to model time series, the strong representing ability of graph neural networks make these three models behave better than other baseline methods. 
It is noteworthy that MTHetGNN outperforms the strong graph-based baseline TEGNN, especially on datasets containing plenty variables, indicating the strong information aggregation capabilities heterogeneous graph networks shows under the same neural network depth. 
This is partly because TEGNN model captures the causality of multivariate time series while MTHetGNN focus on heterogeneity. 
Measuring transfer entropy on the whole time series makes the measurement of causality more accurate. 
However, macroscopic observations will filter out the fluctuations in a single time series segment, thus transfer entropy matrix cannot fully represent the relationship between variables in different time segments. 
Considering this, MTHetGNN not only takes the static relations among time series into account, but considers the dynamic correlations shown in a shorter time segment, fully exploiting the heterogeneity of time series. Detailed analysis are shown in following sections.

\begin{figure}[t]
\centering    
\setlength{\abovecaptionskip}{-0.1cm}
\includegraphics[scale=0.14]{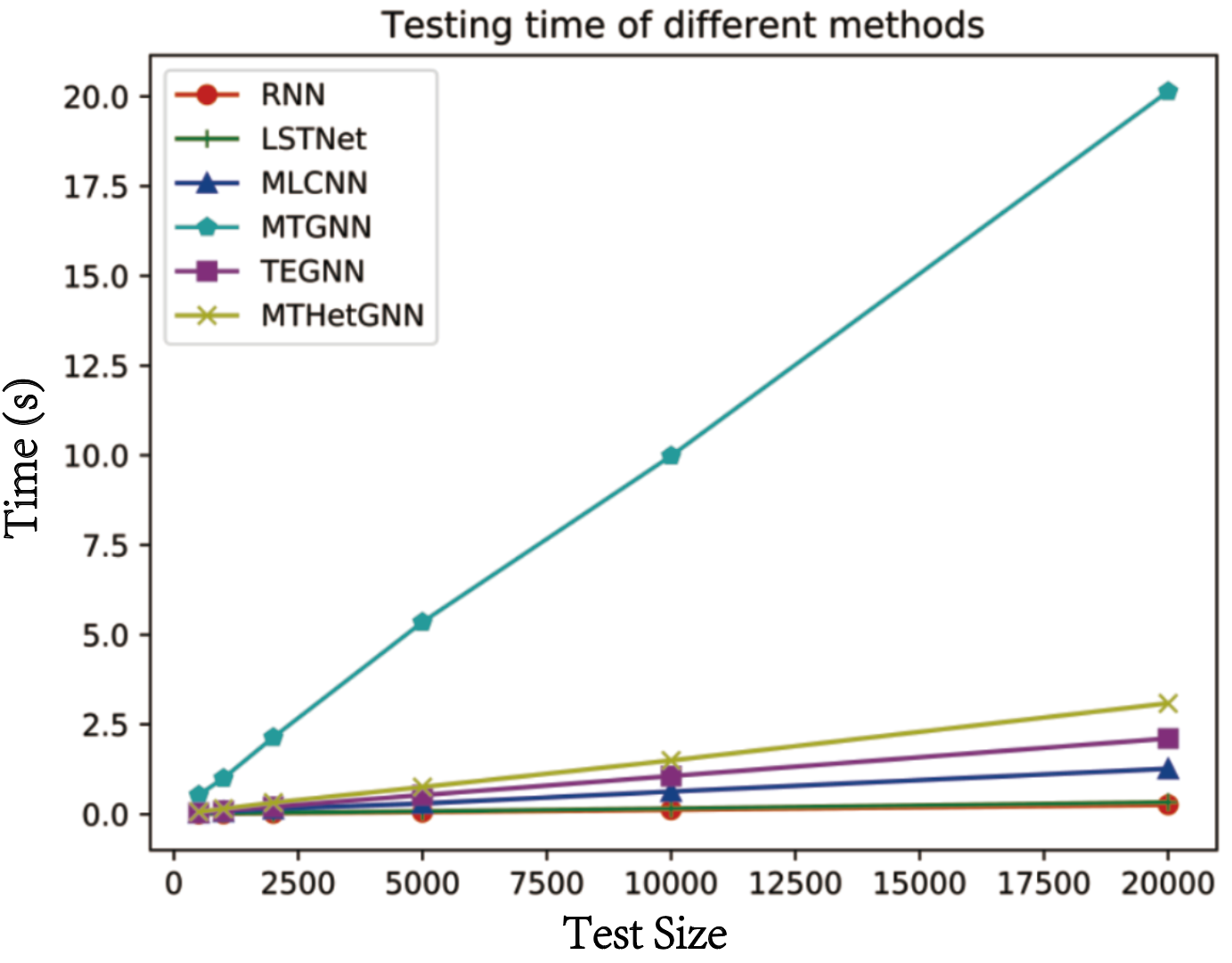} 
\caption{Running time for testing process for all methods on Exchange-Rate dataset when horizon is 3.}
\label{fig:compare_fig1}
\vspace{-0.1cm}
\end{figure}

\subsection{Efficiency}

MTHetGNN has three graph embedding paths, which increases the complexity of the model to a certain extent.
But the node feature matrix generated by temporal embedding module is shared by all the three paths.
And the adjacency matrix $A^{TE}$ and $A^{CO}$ can be calculated in the offline modeling stage, which has little effect on the model complexity.
In order to verify the time complexity of the MTHetGNN model, we record the testing time of the MTHetGNN model and other methods on the Exchange-Rate dataset.
As shown in Figure \ref{fig:compare_fig1}, the MTHetGNN model mines time series multi relations while having relatively high computing efficiency.
Since MTGNN re-extracts features using dilated convolution during each propagation, while other methods in the figure only extract once, the running time of MTGNN is significantly longer than other methods.

\subsection{Ablation Study}
In this subsection, we conduct ablation studies on Exchange-Rate dataset to understand the contributions of heterogeneous graph network in MTHetGNN.
There are two main types of settings, type1, 2, 3 removes the heterogeneous graph part and use one relation extracting way respectively, type4 replaces the attention part with the average operation.
The detailed setting of each variant model is as followed:
\begin{itemize}
    \item type1: Only the Transfer Entropy matrix is used to integrate neighbor information in each layer.
    \item type2: Only the Correlation Coefficient matrix is used to integrate neighbor information in each layer.
    \item type3: Only the Dynamic matrix is used  to integrate neighbor information in each layer.
    \item type4: The MTHetGNN model without attention component, in which the adjacency matrix obtained by three strategies are averaged to a single matrix.
\end{itemize}

\begin{figure}[t]
    \centering
        \includegraphics[scale=0.14]{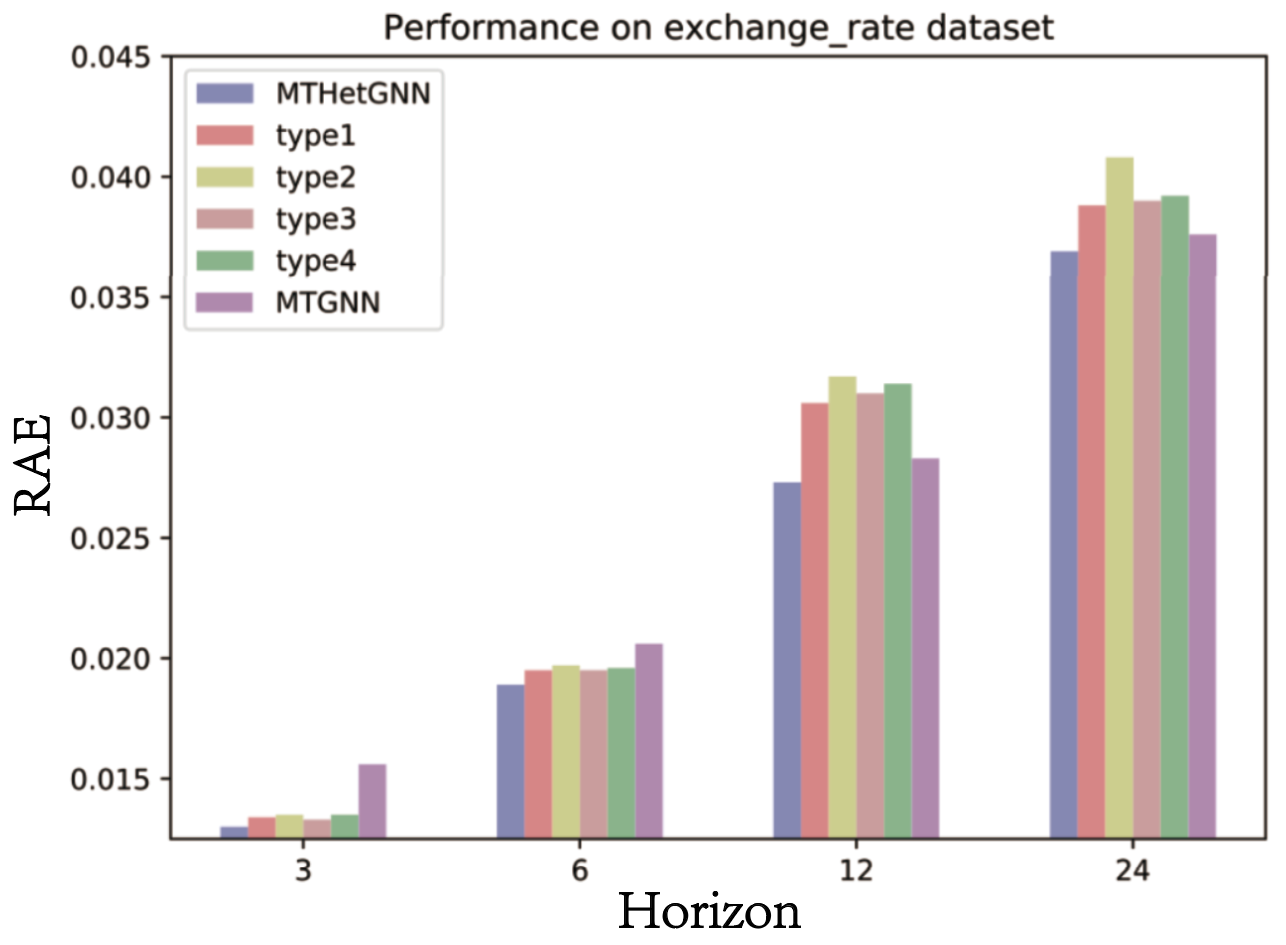}\\
    \centering
    \vspace{-0.4cm}
    \caption{Performance of MTHetGNN and four variants on Exchange-Rate dataset after training 100 epochs. The experiment settings are the same for these methods.}
    \label{fig:compare_fig}
\end{figure}

The results are shown in figure \ref{fig:compare_fig}. We notice that MTHetGNN can model the time series trend more precisely than each variant model, which indicates the effectiveness of both heterogeneous network embedding and attention mechanism in modeling MTS. As is shown, using heterogeneous graph instead of each relation graph raises the \textit{rse} metric of $5.1\%, 7.2\%, 4.5\%$ respectively for type1, 2, 3. It is not surprising given the motivation of using heterogeneous graph. Type1, 2, 3 each only considers relation of the time series in one perspective, while MTHetGNN adopts the concept of heterogeneous graph and fuses the relations in both static and dynamic way. The difference in results between MTHetGNN and type4 indicates the effectiveness of attention mechanism.
By adopting attention mechanism, MTHetGNN can treat each relation graph with different weight, in accordance with the importance of each type of relation altered in training process.

\subsection{Parameter Analysis}
\begin{figure}[t]
    \centering
    \includegraphics[width=0.4\textwidth]{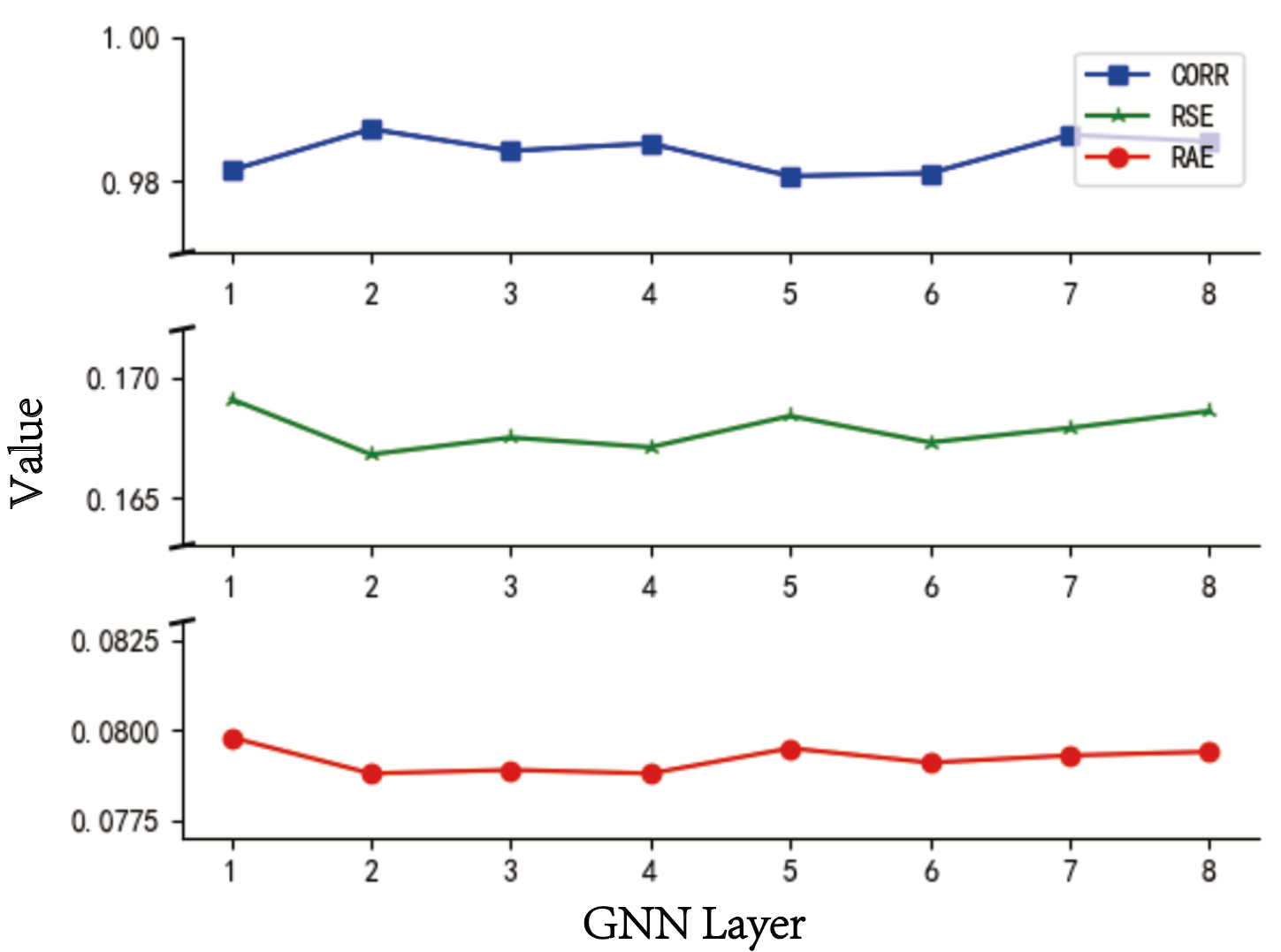}\\
    \centering
    \caption{Performance of MTHetGNN on Solar dataset when horizon is 3. The hidden size of GNN layers is varying while other hyper-parameters remain the same.}
    \vspace{-0.2cm}
    \label{fig:compare_fig2}
\vspace{-0.1cm}
\end{figure}

Meanwhile, we change the network parameters of the heterogeneous graph network and test the performance of the MTHetGNN model under different parameter settings on Solar dataset.
Figure \ref{fig:compare_fig2} shows that the MTHetGNN model does not rely on specific parameters, being relatively not sensitive to parameter changes, showing its effectiveness and stability.

\section{Conclusion}
In this paper, we propose a novel heterogeneous graph embedding based framework (MTHetGNN) for MTS forecasting.
MTHetGNN can exploit rich spatial relation information and temporal features generated by MTS.
Experiments on three real-world datasets show that our model outperforms other baselines in terms of three metrics. 

\section*{Acknowledgments}
This work is supported in part by 
the National Natural Science Foundation of China (No.62002035), the Natural Science Foundation of Chongqing(No.cstc2020jcyj-bshX0034).

\bibliography{mybibfile}

\end{document}